\documentclass[sigconf, screen]{acmart}
\makeatletter
\def\@ACM@checkaffil{
    \if@ACM@instpresent\else
    \ClassWarningNoLine{\@classname}{No institution present for an affiliation}%
    \fi
    \if@ACM@citypresent\else
    \ClassWarningNoLine{\@classname}{No city present for an affiliation}%
    \fi
    \if@ACM@countrypresent\else
        \ClassWarningNoLine{\@classname}{No country present for an affiliation}%
    \fi
}
\makeatother

\usepackage{tabularx} 
\usepackage{booktabs} 
\usepackage[normalem]{ulem} 
\AtBeginDocument{%
  }

\setcopyright{acmlicensed}
\copyrightyear{2018}
\acmYear{2018}
\acmDOI{XXXXXXX.XXXXXXX}

\acmConference[Conference acronym 'XX]{Make sure to enter the correct
  conference title from your rights confirmation emai}{June 03--05,
  2018}{Woodstock, NY}
\acmISBN{978-1-4503-XXXX-X/18/06}

\begin{document}

\title[Evaluating the Explanation Capabilities of LLMs in Conversation Compared to a Human Baseline]{"Is ChatGPT a Better Explainer than My Professor?": Evaluating the Explanation Capabilities of LLMs in Conversation Compared to a Human Baseline}

\author{Grace Li}
\email{gl2676@barnard.edu}
\affiliation{%
  \institution{Barnard College}
}

\author{Milad Alshomary}
\email{ma4608@columbia.edu}
\affiliation{%
  \institution{Columbia University}
}

\author{Smaranda Muresan}
\email{smuresan@barnard.edu}
\affiliation{%
  \institution{Barnard College}
}

\renewcommand{\shortauthors}{Li et al.}
\renewcommand\footnotetextcopyrightpermission[1]{} 

\begin{abstract}
Explanations form the foundation of knowledge sharing and build upon communication principles, social dynamics, and learning theories. We focus specifically on conversational approaches for explanations because the context is highly adaptive and interactive. Our research leverages previous work on explanatory acts, a framework for understanding the different strategies that explainers and explainees employ in a conversation to both explain, understand, and engage with the other party. We use the 5-Levels dataset was constructed from the WIRED YouTube series by Wachsmuth et al., and later annotated by Booshehri et al. with explanatory acts \cite{wachsmuth2022mama}. These annotations provide a framework for understanding how explainers and explainees structure their response when crafting a response. 


With the rise of generative AI in the past year, we hope to better understand the capabilities of Large Language Models (LLMs) and how they can augment expert explainer's capabilities in conversational settings. To achieve this goal, the 5-Levels dataset \footnote{We use Booshehri et al.'s 2023 annotated dataset with explanatory acts.} allows us to audit the ability of LLMs in engaging in explanation dialogues. To evaluate the effectiveness of LLMs in generating explainer responses, we compared 3 different strategies, we asked human annotators to evaluate 3 different strategies: 

\begin{enumerate}
    \item S1: Baseline - human explainer response
    \item S2: GPT4 Standard - GPT explainer response given the previous conversational context
    \item S3: GPT4 w/ EA - GPT explainer response given the previous conversational context and a sequence of explanatory act(s) (EAs) to integrate into its response.
\end{enumerate}

 We found that the GPT generated explainer responses were preferred over the human baseline emphasizing the challenge of effective science communication between experts and everyday people. Additionally, the annotators preferred S2: GPT Standard responses over S2: GPT w/ EA responses mainly due to the concise and succinct responses. For the few times that S3 outperformed S2, annotators noted dimensions of explainee engagement and use of thought-provoking questions as the main reasons for better performance, demonstrating the value in providing explicit instructions for an LLM to follow when generating a response. These results demonstrate the ability of LLMs to generate responses based on sequences of  explanatory acts, allowing for future research to explore the specific contexts and strategies of explanations to improve science communication. Additionally, the results demonstrate the capabilities of LLMs to improve expert explainers' conversational skills and strategies, further emphasizing how interfaces can improve and augment an explainer's abilities. 
\end{abstract}



\begin{teaserfigure}
  \includegraphics[width=\textwidth]{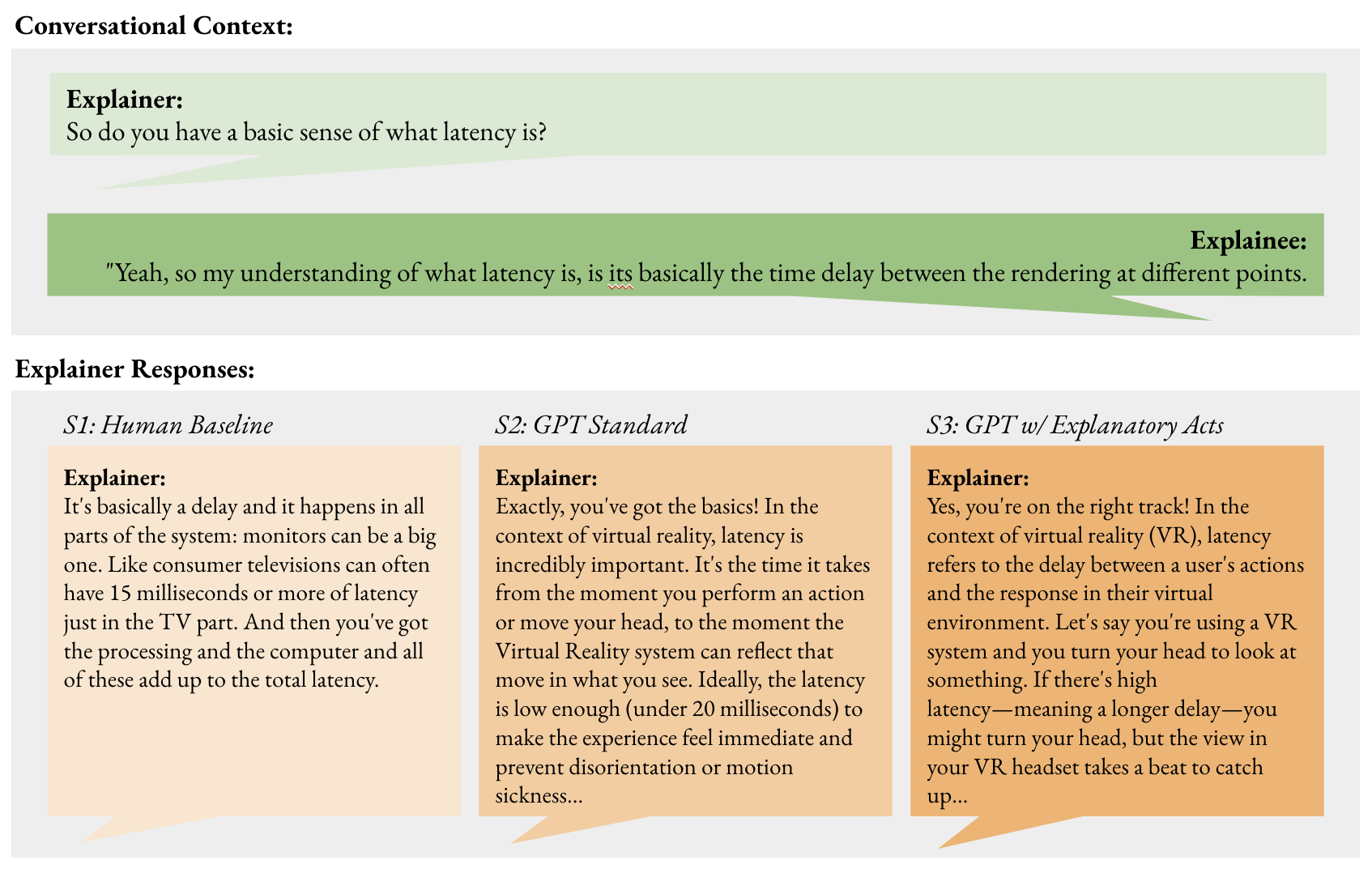}
  \caption{Sample generations from for a given explanation conversation.}
  \label{fig:teaser}
\end{teaserfigure}


\maketitle

\section{Introduction}
\subsection{Background}
Explanations are an important part of science communication because they make science more accessible to the general audience. But it can be hard to bridge the knowledge gap between expert explainers and everyday people who have no prerequisite knowledge of the topic. In this research, we focus specifically on explanation conversations where both the explainer and explainee are engaged in a dialogue to help the explainee understand a concept. These explanation conversations are rich for investigation because the flow of these conversations change and adapt depending on the context and background of the explainer and explainee engaged in the conversation \cite{wachsmuth2022mama}. For example, the method that an explainer would take to explain a concept to a 5-year old will be vastly different from how they would explain the concept to a college student. Various factors such as the explainer and explainee's proficiency and personal interest in the subject area affect how each party will engage in the conversation. This raises the question: \textit{How can explainers tailor their explanation to the explainee's background and proficiency level to increase the explainee's understanding of the topic?}

\subsection{Related Work}
\begin{figure}[!htb]
    \centering
    \includegraphics[width=0.5\textwidth]{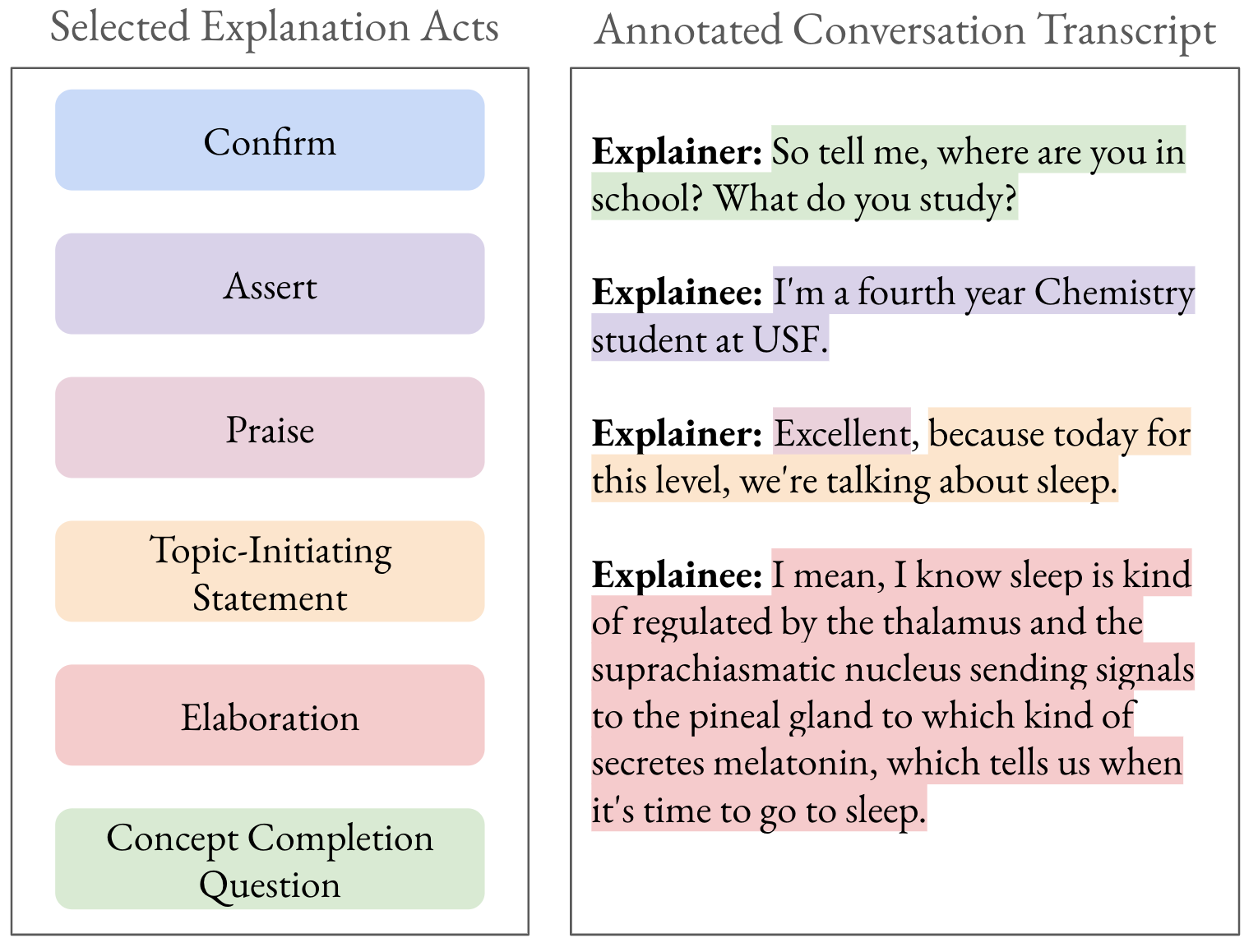}
    \caption{A sample annotated conversation between an explainer and explainee that has been labeled with a subset of the 20 explanatory acts. The figure illustrates the span-level, finegrain annotation framework of Booshehri et al.}
    \label{fig:labeled-dialogue}
\end{figure}

Previous research has focused on creating analytical frameworks to uncover and understand the patterns behind the explanation conversations between explainers and explainees. Booshehri et al. has explored how experts and explainees engage in explanation conversations through an inventory of "explanatory acts," which are categories to characterize the contributions and intentions behind the explainer and explainee's utterances. Booshehri et al. developed 20 explanatory acts for the purpose of fine-grain explanation annotations to increase the understanding in terms of the interaction dynamics between the explainer and explainee. In Figure \ref{fig:labeled-dialogue}, an example of an annotated conversation using Boosherhri's inventory of explanatory acts illustrates how sentences can be broken down into multiple different explanatory acts. By focusing on span-level annotations, Boosherhri's inventory of explanation moves allows for a fine-grain categorization and understanding of the different strategies that explainers and explainees use in their conversations with each other. For example, the explanatory acts include categories like Elaboration, Definition question, Analogy, and more to pinpoint the specific strategies that explainers and explainees employ in a conversation.  

The entire list of explanatory acts is included in the Appendix of the paper. While this research focuses on developing a framework to understand how human explainers and explainees explain topics to each other, there is still a lack of research comparing the effectiveness of human explanations and those generated from Large Language Models (LLMs). 

\subsection{Large Language Models}
The field of communication has also shifted due to the increasing availability of LLMs, which have raised concerns about the effectiveness of LLM-generated explanations and the reliability of the generated text. While these LLMs have been trained on vast amounts of data that have been sourced from the internet, not much is known about whether these models have internalized the ability to model human-like explanations. Furthermore, little research has been done to evaluate language models on their abilities to engage in explanatory conversations as the role of an explainer. Our research aims to shed light on two areas: first, how effectively LLMs are able to generate explainer responses and second, whether an LLM can formulate a response based on a given sequence of explanation moves. These two areas will help better understand how LLM generated responses compare to human responses and can provide insights into how to develop interactive explanation systems and how LLMs can better augment human explanation capabilities to improve the quality of human-explainer responses. Additionally, by evaluating whether LLMs are able to formulate responses based on explanatory moves, can further the field of explainable AI (XAI) systems in understanding how well LLMs are able to model explanation responses based on these explanation frameworks.

We design a study performs a side-by-side comparison on human expert responses to different LLM-generated responses to better understand the effectiveness of LLM explanations in a science explanation dialogues. We hypothesize that because LLMs are trained on large sources of data, their explanation qualities might implicitly model human explanations, but would require additional scaffolding to ensure consistency in maintaining engagement with the explainee. 

\section{Methods}
\subsection{Data}
In this study, we use Booshehri's annotated WIRED magazine's 5 Levels of Explanation Youtube video dataset from 3 different annotators with explanatory act labels from research's proposed inventory. WIRED magazine's "5 Level Video Series," contains conversations between one expert with 5 different people, each at a different level of proficiency in the topic: a child, a teenager, an undergraduate student, a graduate student, and a colleague. This dataset is most suitable for our specific use case because it illustrates staged conversations between an expert explainer and an explainee that are filmed in a studio environment. The staged environment allows for the explanation to be distilled down to its core components without the noise that might occur from in-the-field explanations. These staged conversations allow for both the explainer and explainee to succinctly engage with each other to understand a certain topic. 

For this study, we specifically focus on conversations between an expert explainer and a college-level explainee for the purpose to standardizing the evaluation metric. We choose college-level explainees because we found these conversations yielded the best balance of depth and technical nuance for a concept. When explaining to a child and high school student, the explainer over simplified the topics. Alternatively, with graduate students and other colleagues, the explainer dove straight into the technical mechanics of the topics without any preliminary topic introduction. With college students, however, the explainer often provided enough context to the topic that a general audience member could understand while also providing additional technical depth. Additionally, we focused on STEM topics to align the research with the goal of improving science communication. There were 11 topics (virtual reality, sleep, nano-technology, machine learning, lasers, hacking, gravity, dimensions, connectomes, blockchain, and blackholes). 

\subsection{Study Design}
\begin{figure}[ht]
    \centering
    \includegraphics[width=0.5\textwidth]{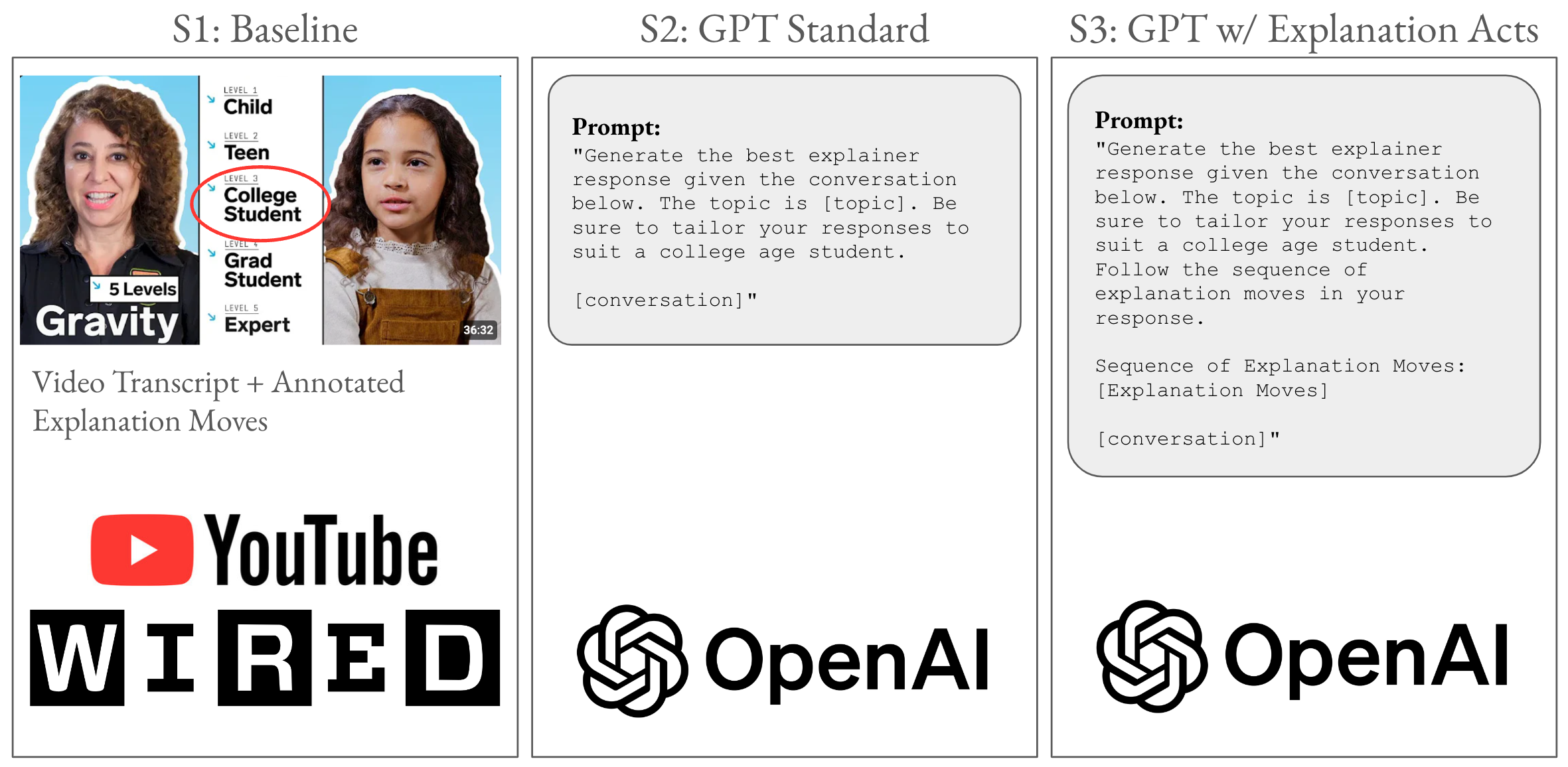}
    \caption{The three different study conditions.}
    \label{fig:study-conditions}
\end{figure}

The purpose of the study was to evaluate three different methods to generate explainer responses to an explanation conversation. The first approach, S1, is the Baseline approach that uses the human explainer's actual response from the 5-Levels dataset to structure the response. The second approach, S2, is the Standard Prompting approach that provides the previous conversation context to OpenAI's GPT4 and asks the LLM to generate an explainer's response \cite{achiam2023gpt}. The third approach, S3, is the Prompting with Explanation Acts (EAs) approach that provides the previous conversation context and the sequence of observed explanatory acts from the annotated 5-Levels dataset as an outline for the LLM to follow for GPT to follow as it's generating it's response \cite{achiam2023gpt}. Figure \ref{fig:study-conditions} illustrates the different prompting strategies.

To evaluate the different strategies of generating explainer responses, we used the 5-Levels dataset to create the conversational context for each explainer response. We manually parsed the 5-Levels dataset to incrementally concatenate pairs of explainer and explainee utterances together to build out an entire conversation, ensuring that every sequence ends on an explainee utterance. Ending on an explainee utterance is important because it allows the explainer the ability to directly or indirectly respond to the explainee's last utterance. In this manner, for every explainee utterance in a conversation, we generate a corresponding explainer utterance given the two different prompting strategies illustrated in Figure \ref{fig:study-conditions}. In the study, we randomized the order that each response condition was displayed, changing the ordering of each condition. Each user experienced the same sequence of randomized response labels.

\subsection{Evaluation Criteria}
\begin{figure}[ht]
    \centering
    \includegraphics[width=0.5\textwidth]{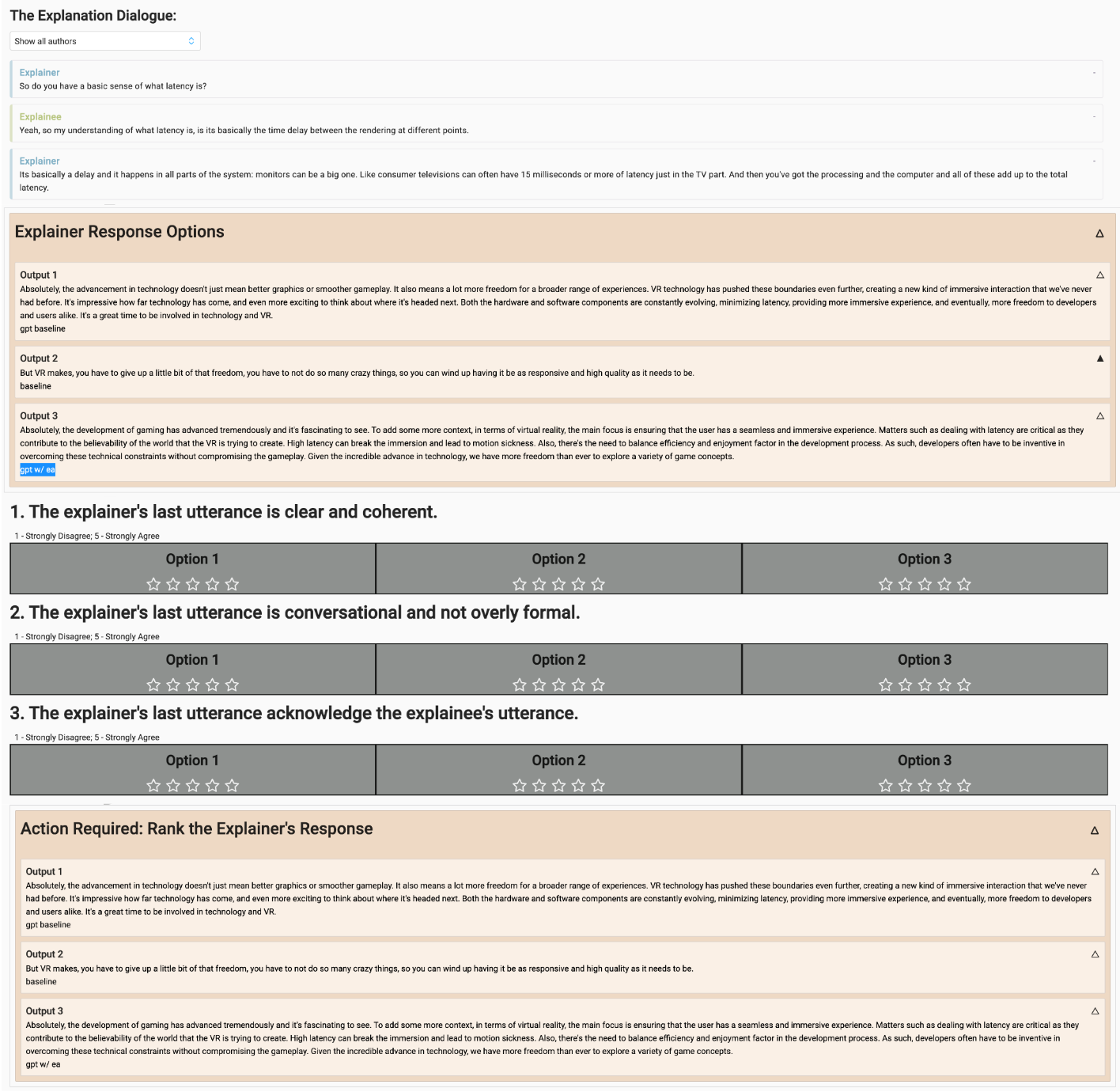}
    \caption{A sample image of the annotation interface with some of the rating questions omitted. The interface contains 3 columns of 8 rows with 5-star ratings in each of the columns to evaluate the explainer responses on 8-dimensions: coherence, conciseness, conversational nature, appropriateness, acknowledgement, active guidance, engagement, and depth or expansiveness}
    \label{fig:annotation-interface}
\end{figure}
We designed a custom evaluation interface in Label Studio\footnote{Label Studio: https://labelstud.io/} that scored each explainer response on 8 different dimensions on a Likert rating scale of 1-5 (ranging from Strongly Disagree to Strongly Agree) and one ranking question that evaluated the different explainer responses against each other. The 8 dimensions that each explainer response was evaluated on included:
\begin{enumerate}
    \item Coherence: The explainer's last utterance is clear and coherent.
    \item Concise: The explainer's last utterance is concise.
    \item Conversational: The explainer's last utterance is conversational and not overly formal.
    \item Acknowledgement: The explainer's last utterance acknowledges the explainee's utterance.
    \item Appropriate: The explainer's last utterance responds appropriately to the explainee's utterance.
    \item Deepens or Expands: In the context of the entire conversation, the explainer's last utterance deepens or expands the conversation.
    \item Actively Guidance:  In the context of the entire conversation, the explainer's last utterance actively guides the course of the conversation.
    \item Engagement of Explainee: In the context of the entire conversation, the explainers last utterance engages the explainee in the conversation.
\end{enumerate}

These dimensions were designed based on Li et al.'s questionnaire design to evaluate a chatbot's responses in the setting of evaluating the effectiveness of different chatbots to assist English language learners to enhance their conversational skills \cite{li2023curriculum}. We used the language and content quality dimensions from Li et al.'s questionnaire design to inform parts of our evaluation criteria and included additional questions to further probe the question of the efficacy of the explainer responses as it relates to previous findings in the field of effective explanations.

The rating section allowed annotators to rank the different outputs against each other through a drag and drop interface. This provided insights into how well certain conditions performed in relation to each other. In addition to the rating scale, the annotators were also asked to provide a rationale for their ranking. The rating system does not allow for ties, so each of the three explainer responses had to be assigned a unique value from 1-3. The evaluation interface was designed in LabelStudio. Figure \ref{fig:annotation-interface} illustrates a sample interface that has been shortened.

\subsection{Participant Recruitment}
We recruited participants from a platform called Upwork, a professional crowd working platform. We hired 3 annotators all with a 100\% job success rate on the platform. Each annotator labeled 104 tasks, each task included evaluating 3 different explainer responses on 8 different dimensions and ranking the 3 responses against each other and providing a response for the ranking, which resulted in 26 questions for each task. We paid each annotator \$135 for around 7-10 hours of work. All hired annotators signed a consent form and were on-boarded onto the annotation platform, LabelStudio, where they received detailed instructions for how to complete the annotations. 

\section{Results}
We calculated the inter-annotator agreement score for each of the two sections: 8-dimension rating section and the ranking section. We used Krippendorff’s alpha to evaluate the inter annotator agreement on each of the 8-dimensions.  We then used Kendall's Tau to calculate the pairwise inter-annotator agreement for each task's rankings. We found that Annotators 21549 and Annotator 21551 had a pairwise agreement of 0.42–illustrating a moderate agreement in rankings. All three annotators only had an inter-annotator agreement score of 0.167. Demonstrates how this annotation task is highly variable due to each annotator's own specific preferences for engaging in explanation conversations. 

\begin{center}
\begin{tabular}{||c c c c||} 
 \hline
  & Rank 1 & Rank 2 & Rank 3  \\ [0.5ex] 
 \hline\hline
 S1: Baseline & 18\% & 22\% & 59\% \\
   [1ex] 
 \hline
 \textbf{S2: GPT Standard} & \textbf{49\%} & 34\% & 17\% \\
   [1ex] 
 \hline
 S3: GPT w/ EA  & 33\% & 44\% & 23\% \\
  [1ex] 
 \hline
\end{tabular}
\captionof{table}{Percent distribution of S1, S2, and S3 explainer results for each ranking. Results were evaluated over all 312 annotator labels.}
\label{results-table}
\end{center}

As seen in Table \ref{results-table}, S2: GPT Standard resulted in the 49\% of the Rank 1 results, demonstrating that it outperforms S1 and S3. Comparatively, S1: Baseline, performs the worst with over 59\% of it's outputs being ranked last, Rank 3 out of 3 possible choices. To better understand the differences in ranking between S2: GPT Standard and S3: GPT w/ EA, Table \ref{detailed-results-table} illustrates more detailed percentages and evaluations on when each condition outperforms the other. 

According to Table \ref{detailed-results-table}, 35\% of all annotated tasks were labeled with S2: GPT Standard in the Rank 1 position and S3: GPT w/ EA in the Rank 2 position. The most common rationale for this ranking was because the S2 strategy was "a little too long," "overly wordy," "long winded" and "over-explains in several areas and is longer than necessary" according to annotators. On average, S3 - GPT w/ EA responses were 10 words longer than S2 - GPT Standard responses. 

Alternatively, only 24\% of all annotated tasks labeled S3:GPT w/ EA in Rank 1 and S2: GPT Standard in Rank 2. The rational that many annotators wrote included responses such as "actively guides the conversation," "engaged the explainee with a followup question," and "asks a thought provoking question prompting deeper conversation." 

\begin{center}
\begin{tabular}{||c c | c||} 
 \hline
 Rank 1 & Rank 2 & \textbf{Percentage}\\ [0.5ex] 
 \hline
 \hline
 S2: GPT Standard & S3: GPT w/ EA & \textbf{35\%} \\
   [1ex] 
 \hline
 S3: GPT w/ EA& S2: GPT Standard & \textbf{24\%} \\
   [1ex] 
 \hline
\end{tabular}
\captionof{table}{Percent distribution of the when S2 and S3 rank first and second out of all 312 annotated occurrences.}
\label{detailed-results-table}
\end{center}

\section{Discussion and Future Work}
This study further demonstrates that more work needs to be done to help experts bridge the knowledge gap between themselves and their audiences. While LLM-generated responses have been shown to perform better than the baseline human responses, these findings cannot be used to advocate for LLMs to replace the function of expert explainers. Instead, this research demonstrates how LLMs are able to augment expert explainer's capabilities by offering real-time support in tailoring more effective explanation for a given audience. Additionally, based on the qualitative results from the annotators's responses, one of the main reasons that S2: GPT Standard outperformed S3: GPT w/ EA was due to it's conciseness, with an average of 10 fewer words per response. This demonstrates that being concise is important in not overwhelming the explainee with information and how carefully planning and segmenting an explanation into manageable chunks is important for information communication and retention. One area that S3: GPT w/ EA performed better than S2: GPT Standard was in the structure of the responses, specifically in generating engaging followup or thought-provoking conversations. This demonstrates that the instances where GPT was explicitly prompted to include a question such as a concept completion question or test understanding question, annotators felt more engaged and guided by the conversation. Given that LLM is following instructions regarding what explanation moves to follow, we can argue that prompting LLMs with explanation moves helps avoiding redundant acts and dull conversation and force them to use more novel moves.


Current research that evaluates the efficacy of chatbot interfaces for helping students understand complex topics is an ongoing area of research. Most of the research focuses on the explainee's experience with LLM-powered chatbot interfaces and works on helping the explainee frame or scope their questions to improve the quality of the outputs that an LLM gives them. In parallel to these ongoing research projects, more work needs to be done on evaluating how LLMs can augment the capabilities of expert explainers'. How can interface design best support human explainers? Our research illustrates that LLMs are able to generate a response given a sequences of explanatory acts which demonstrates that if given the most effective strategy to respond, an LLM will be able to formulate a response following that explanation structure. This allows for further research into formulations of effective explanation strategies, distilling them into a sequence of explanation acts that an LLM can execute. Additionally, as seen by the low inter-annotator agreement, future research can be conducted in designing system to aid in automatic personalization of explanations, conversation structures and styles to improve the experience regardless of personal preferences--allowing for an adaptable experience based on the explainee. 







\bibliographystyle{ACM-Reference-Format}
\bibliography{sample-base}


\begin{thebibliography}{3}


\ifx \showCODEN    \undefined \def \showCODEN     #1{\unskip}     \fi
\ifx \showDOI      \undefined \def \showDOI       #1{#1}\fi
\ifx \showISBNx    \undefined \def \showISBNx     #1{\unskip}     \fi
\ifx \showISBNxiii \undefined \def \showISBNxiii  #1{\unskip}     \fi
\ifx \showISSN     \undefined \def \showISSN      #1{\unskip}     \fi
\ifx \showLCCN     \undefined \def \showLCCN      #1{\unskip}     \fi
\ifx \shownote     \undefined \def \shownote      #1{#1}          \fi
\ifx \showarticletitle \undefined \def \showarticletitle #1{#1}   \fi
\ifx \showURL      \undefined \def \showURL       {\relax}        \fi
\providecommand\bibfield[2]{#2}
\providecommand\bibinfo[2]{#2}
\providecommand\natexlab[1]{#1}
\providecommand\showeprint[2][]{arXiv:#2}

\bibitem[Achiam et~al\mbox{.}(2023)]%
        {achiam2023gpt}
\bibfield{author}{\bibinfo{person}{Josh Achiam}, \bibinfo{person}{Steven Adler}, \bibinfo{person}{Sandhini Agarwal}, \bibinfo{person}{Lama Ahmad}, \bibinfo{person}{Ilge Akkaya}, \bibinfo{person}{Florencia~Leoni Aleman}, \bibinfo{person}{Diogo Almeida}, \bibinfo{person}{Janko Altenschmidt}, \bibinfo{person}{Sam Altman}, \bibinfo{person}{Shyamal Anadkat}, {et~al\mbox{.}}} \bibinfo{year}{2023}\natexlab{}.
\newblock \showarticletitle{Gpt-4 technical report}.
\newblock \bibinfo{journal}{\emph{arXiv preprint arXiv:2303.08774}} (\bibinfo{year}{2023}).
\newblock


\bibitem[Li et~al\mbox{.}(2023)]%
        {li2023curriculum}
\bibfield{author}{\bibinfo{person}{Yu Li}, \bibinfo{person}{Shang Qu}, \bibinfo{person}{Jili Shen}, \bibinfo{person}{Shangchao Min}, {and} \bibinfo{person}{Zhou Yu}.} \bibinfo{year}{2023}\natexlab{}.
\newblock \showarticletitle{Curriculum-Driven Edubot: A Framework for Developing Language Learning Chatbots Through Synthesizing Conversational Data}.
\newblock \bibinfo{journal}{\emph{arXiv preprint arXiv:2309.16804}} (\bibinfo{year}{2023}).
\newblock


\bibitem[Wachsmuth and Alshomary(2022)]%
        {wachsmuth2022mama}
\bibfield{author}{\bibinfo{person}{Henning Wachsmuth} {and} \bibinfo{person}{Milad Alshomary}.} \bibinfo{year}{2022}\natexlab{}.
\newblock \showarticletitle{" Mama Always Had a Way of Explaining Things So I Could Understand'': A Dialogue Corpus for Learning to Construct Explanations}.
\newblock \bibinfo{journal}{\emph{arXiv preprint arXiv:2209.02508}} (\bibinfo{year}{2022}).
\newblock


\end{thebibliography}

\appendix
\section{Inventory of Explanatory Acts}
\begin{figure}[bp!]
    \centering
    \includegraphics[width=0.5\textwidth]{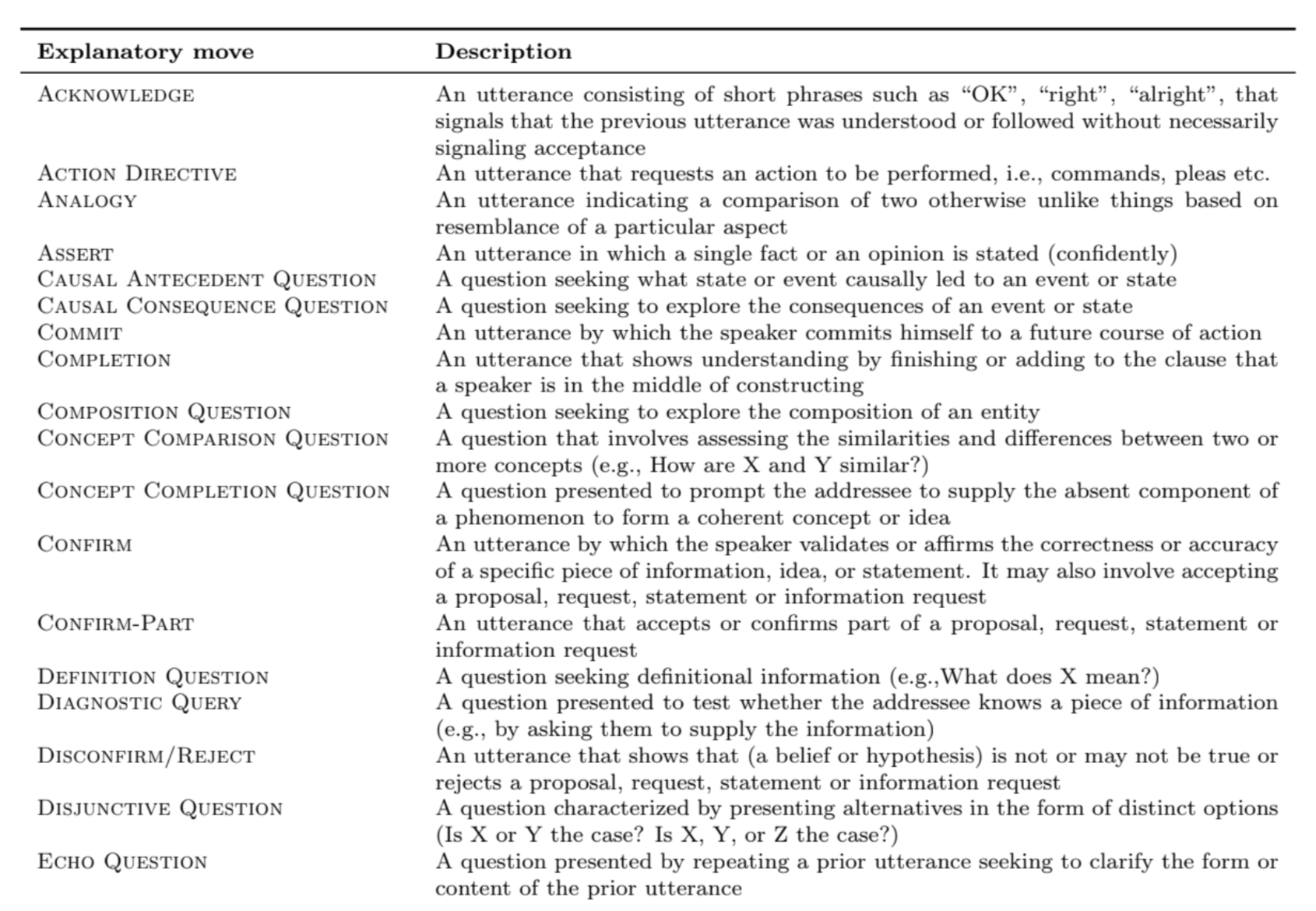}
    \includegraphics[width=0.5\textwidth]{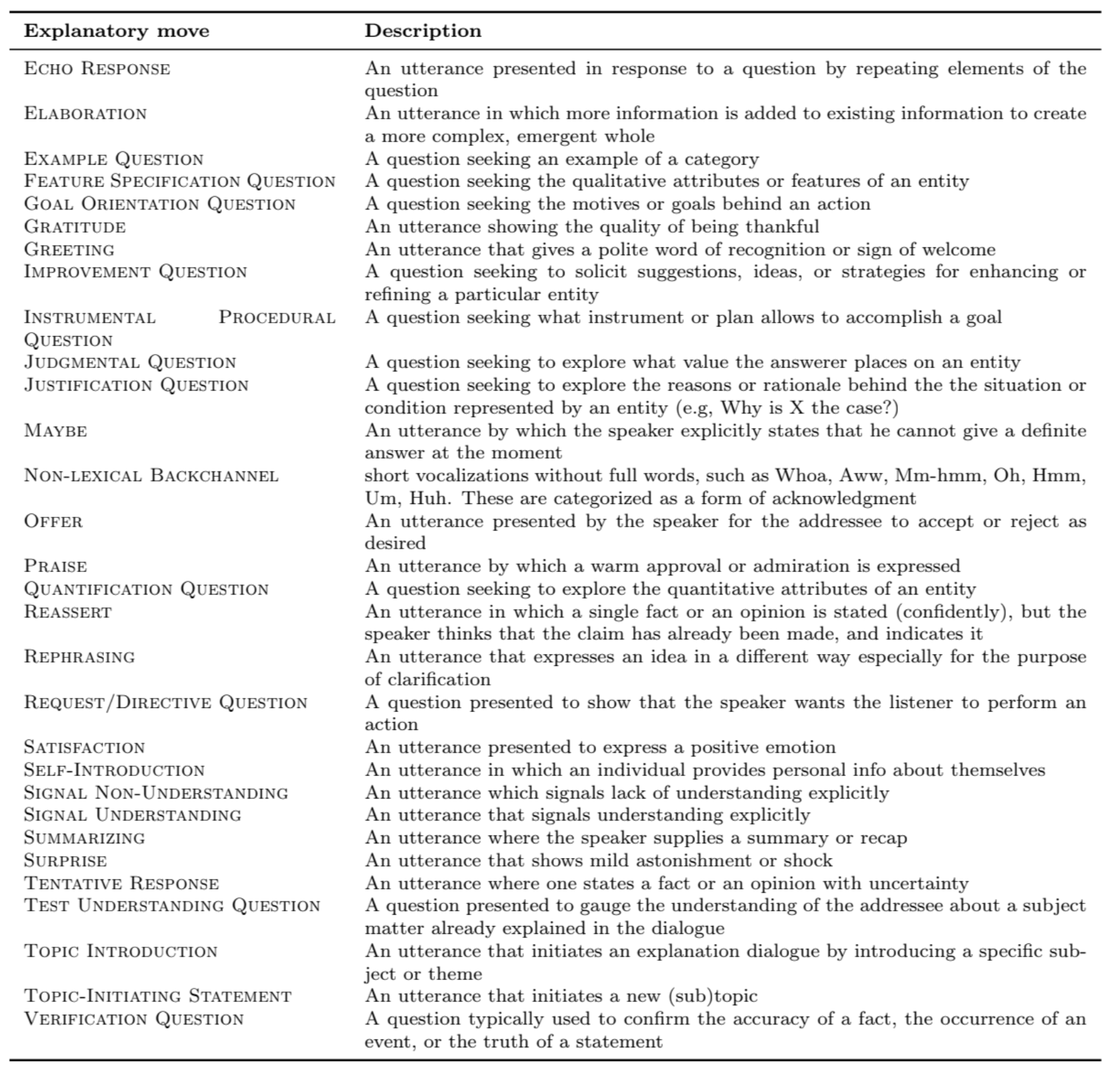}
    \caption{List of explanatory moves in our proposed annotation scheme along with their descriptions, arranged in alphabetical order}
    \label{fig:explanatory-acts}
\end{figure}

\end{document}